\documentclass[letterpaper, 10 pt, conference]{ieeeconf}  

\IEEEoverridecommandlockouts                              

\usepackage{graphics} 
\usepackage{epsfig} 
\usepackage{mathptmx} 
\usepackage{times} 
\usepackage{amsmath} 
\usepackage{amssymb}  
\usepackage{url}
\usepackage{xcolor}
\usepackage{hyperref}
\usepackage{subcaption}
\usepackage{float}

\usepackage{booktabs}
\usepackage{gensymb}

\DeclareMathOperator*{\argmin}{arg\,min}

\title{\LARGE \bf
Generalizable Pose Estimation Using Implicit Scene Representations
}

\author{Vaibhav Saxena$^{1,\dagger}$, Kamal Rahimi Malekshan$^{2}$, Linh Tran$^{2,\ddagger, *}$ and Yotto Koga$^{2,*}$ 
\thanks{$^{1}$ Correspondence to: \url{vsaxena33@gatech.edu} (Georgia Institute of Technology)}
\thanks{$^{2}$ \url{{yotto.koga, kamal.malekshan}@autodesk.com} (Autodesk Research)}
\thanks{$\dagger$ Work done while interning at Autodesk Research.}
\thanks{$\ddagger$ \url{mail@linht.com} (work done while at Autodesk Research.)}
\thanks{$*$ Equal supervision.}
\thanks{Project website: \href{https://sites.google.com/view/generalizable-pose-estimation/}{sites.google.com/view/generalizable-pose-estimation}}
}

\begin{document}

\maketitle
\thispagestyle{empty}
\pagestyle{empty}

%
\begin{abstract}
6-DoF pose estimation is an essential component of robotic manipulation pipelines. However, it usually suffers from a lack of generalization to new instances and object types. Most widely used methods learn to infer the object pose in a discriminative setup where the model filters useful information to infer the exact pose of the object. While such methods offer accurate poses, the model does not store enough information to generalize to new objects. In this work, we address the generalization capability of pose estimation using models that contain enough information about the object to render it in different poses. We follow the line of work that inverts neural renderers to infer the pose. We propose i-$\sigma$SRN to maximize the information flowing from the input pose to the rendered scene and invert them to infer the pose given an input image. Specifically, we extend Scene Representation Networks (SRNs) by incorporating a separate network for density estimation and introduce a new way of obtaining a weighted scene representation. We investigate several ways of initial pose estimates and losses for the neural renderer. Our final evaluation shows a significant improvement in inference performance and speed compared to existing approaches.
\end{abstract}

\section{Introduction}
\label{sec:intro}

Six degrees of freedom (6 DoF) pose estimation is the task of detecting the pose of an object in 3D space, which includes its location and orientation. Pose estimation is a crucial part of robotic grasping and manipulation in various domains, such as manufacturing and assembly~(\cite{litvak2019learning,chen2020repetitive,rocha2014object,choi2012voting}), healthcare~(\cite{chen2018patient,obdrvzalek2012accuracy}), and households~(\cite{tremblay2018deep}).
However, existing methods for pose estimation are limited in their applications. The majority of approaches can only be used for a specific object or for categories (\cite{huttenlocher1993comparing,hinterstoisser2011gradient,xiang2018posecnn,wang2019densefusion,li2019cdpn,peng2019pvnet,he2020pvn3d}) that are similar to the ones in the training data. Recent works attempt to generalize object pose estimation to unseen objects (\cite{he2022fs6d,liu2022gen6d,sun2022onepose}). However, they require high-quality 3D models, additional depth maps, and segmentation masks at test time. These requirements limit these existing pose estimators for real-world applications.

Learning implicit representations of 3D scenes has enabled high-fidelity renderings of scenes \cite{park2019deepsdf,sitzmann2019scene,mescheder2019occupancy,mildenhall2020nerf}, object compression (\cite{tang2020deep,dupont2021coin}),  and scene completion~\cite{sitzmann2020implicit}. It has also opened up novel research directions in robot navigation \cite{adamkiewicz2022nerfnav} and manipulation \cite{simeonov2022ndf}. In contrast to explicit scene representations, implicit representations incorporate 3D coordinates as input to a deep neural network enabling resolution-free representation for all topologies. A recent work, iNerf~\cite{yen2021inerf}, leverages neural rendering and explores pixelNeRF for camera pose optimization. Although iNerf shows promising results, the computational (pre-training a deep neural network) and input (pose and scene render) requirements have impeded its usability for real-world applications.

\begin{figure}[t]
     \centering
     \begin{subfigure}[b]{0.2\linewidth}
         \centering
         \includegraphics[width=\linewidth]{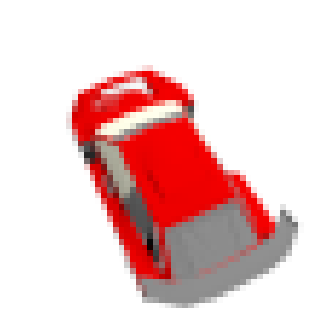}
         \label{fig:conv_D_a}
     \end{subfigure}
     \vrule
     \hfill
     \begin{subfigure}[b]{0.2\linewidth}
         \centering
         \includegraphics[width=\linewidth]{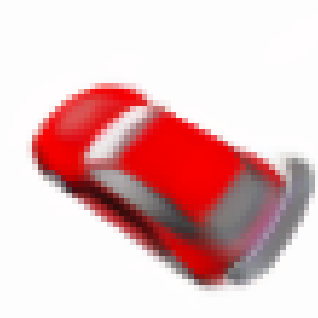}
         \label{fig:conv_D_b}
     \end{subfigure}
     \hfill
     \begin{subfigure}[b]{0.2\linewidth}
         \centering
         \includegraphics[width=\linewidth]{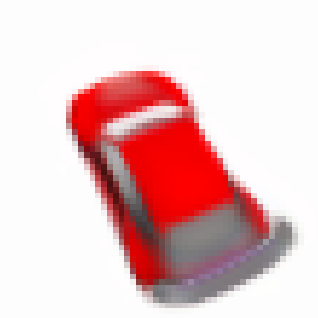}
         \label{fig:conv_D_c}
     \end{subfigure}
     \begin{subfigure}[b]{0.2\linewidth}
         \centering
         \includegraphics[width=\linewidth]{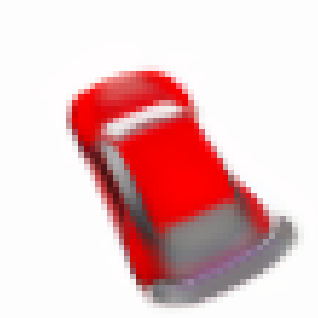}
         \label{fig:conv_D_d}
     \end{subfigure}
     \begin{subfigure}[b]{0.2\linewidth}
         \centering
         \includegraphics[width=\linewidth]{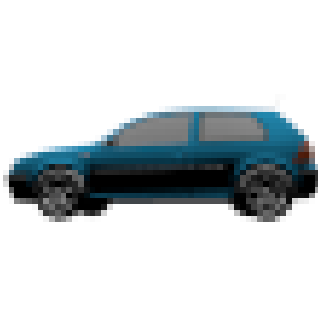}
         \renewcommand{\thesubfigure}{a}%
         \caption*{Target}
         \label{fig:conv_E_a}
     \end{subfigure}
     \vrule
     \hfill
     \begin{subfigure}[b]{0.2\linewidth}
         \centering
         \includegraphics[width=\linewidth]{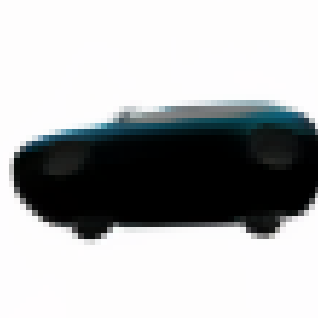}
         \renewcommand{\thesubfigure}{b}%
         \caption*{Initial pose}
         \label{fig:conv_E_b}
     \end{subfigure}
     \hfill
     \begin{subfigure}[b]{0.2\linewidth}
         \centering
         \includegraphics[width=\linewidth]{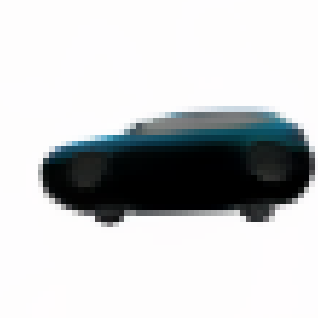}
         \renewcommand{\thesubfigure}{c}%
         \caption*{100 steps}
         \label{fig:conv_E_c}
     \end{subfigure}
     \hfill
     \begin{subfigure}[b]{0.2\linewidth}
         \centering
         \includegraphics[width=\linewidth]{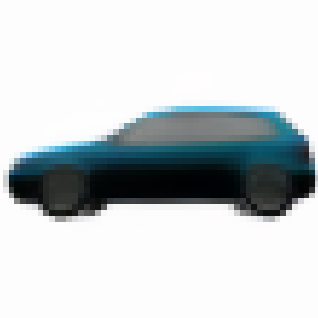}
         \renewcommand{\thesubfigure}{d}%
         \caption*{300 Steps}
         \label{fig:conv_E_d}
     \end{subfigure}
    \caption{\textbf{Pose estimation using i-$\sigma$SRN.} Given a query image, that we treat as the target render, we iteratively refine the pose estimate for 300 steps until our output render closely matches the query image.
    }
    \label{fig:convergence_All}
    \vspace{-0.5cm}
\end{figure}

In this paper, we propose i-$\sigma$SRN, a novel framework for 6-DoF pose estimation that computes poses by inverting an implicit scene representation model trained to render views from arbitrary poses.
We leverage and \textit{extend} the Scene Representation Network (SRN)~\cite{sitzmann2019scene}, a 3D structure-aware scene representation model capable of generalizing to novel object instances using a hypernetwork parameterization, for accurate pose estimation. The main advantage of our pose estimator is that it only requires \textbf{simple inputs} and is \textbf{generalizable}. These simple inputs include RGB images, camera intrinsics, and pose information during training. During pose inference, it only requires RGB images. In contrast to iNerf, our approach does not require a source rendered RGB image but rather only an initial pose estimate.
While classical methods for pose estimation utilize RGB images and depth maps, they are usually impacted by changes in object materials, reflectance, and lighting conditions. Neural rendering approaches implicitly model lighting and reflectance, and thus are more robust to their influence. Our approach is also generalizable as it can be applied to an arbitrary object with minimal additional training (two-shot generalization). 
When generalizing to an unseen object, the estimator only needs a few reference images of the object under known camera poses for training.

We summarize our contributions below: 
\begin{figure*}[t!]
    \centering
    \subcaptionbox{Phase 1: Training the neural renderer $\sigma$SRN}{
        \centering
        \includegraphics[width=0.49\linewidth]{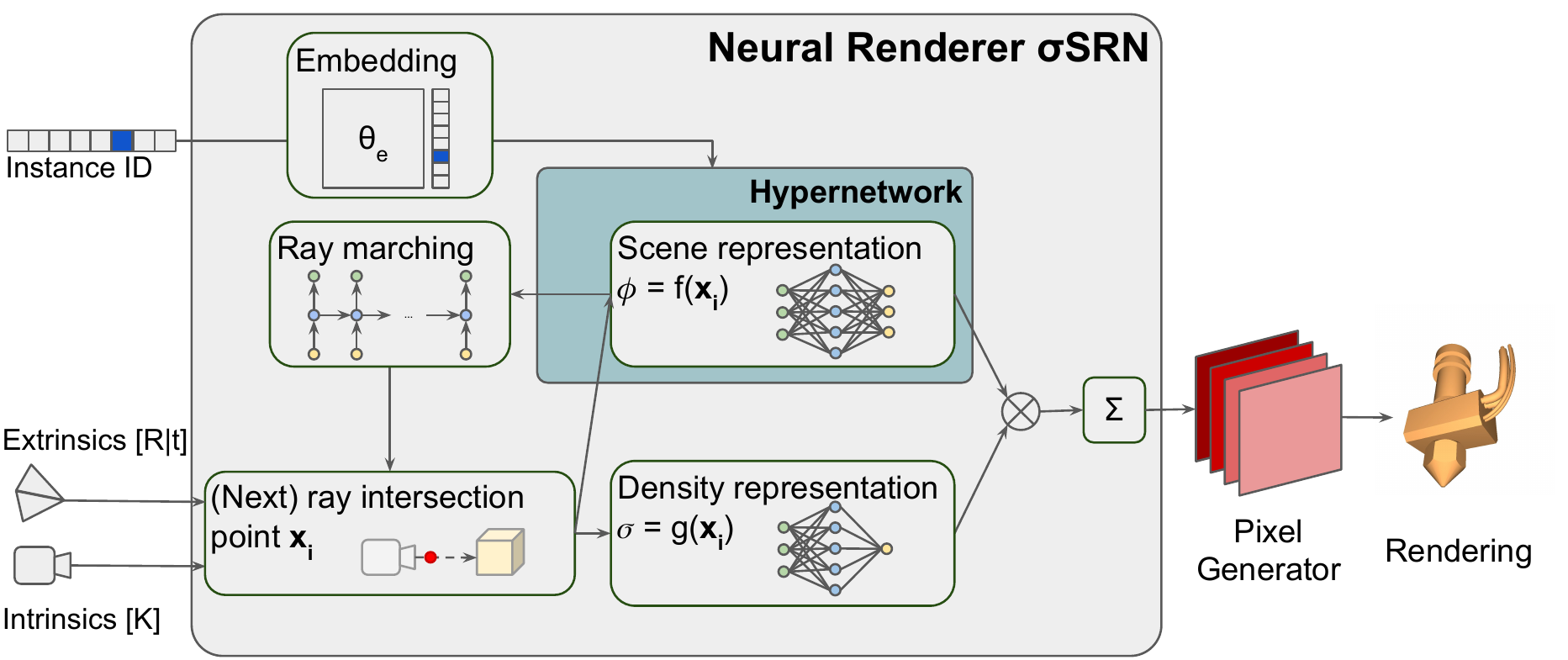}
    }
    \subcaptionbox{Phase 2: Pose estimation by inverting $\sigma$SRN}{
        \centering
        \includegraphics[width=0.46\linewidth]{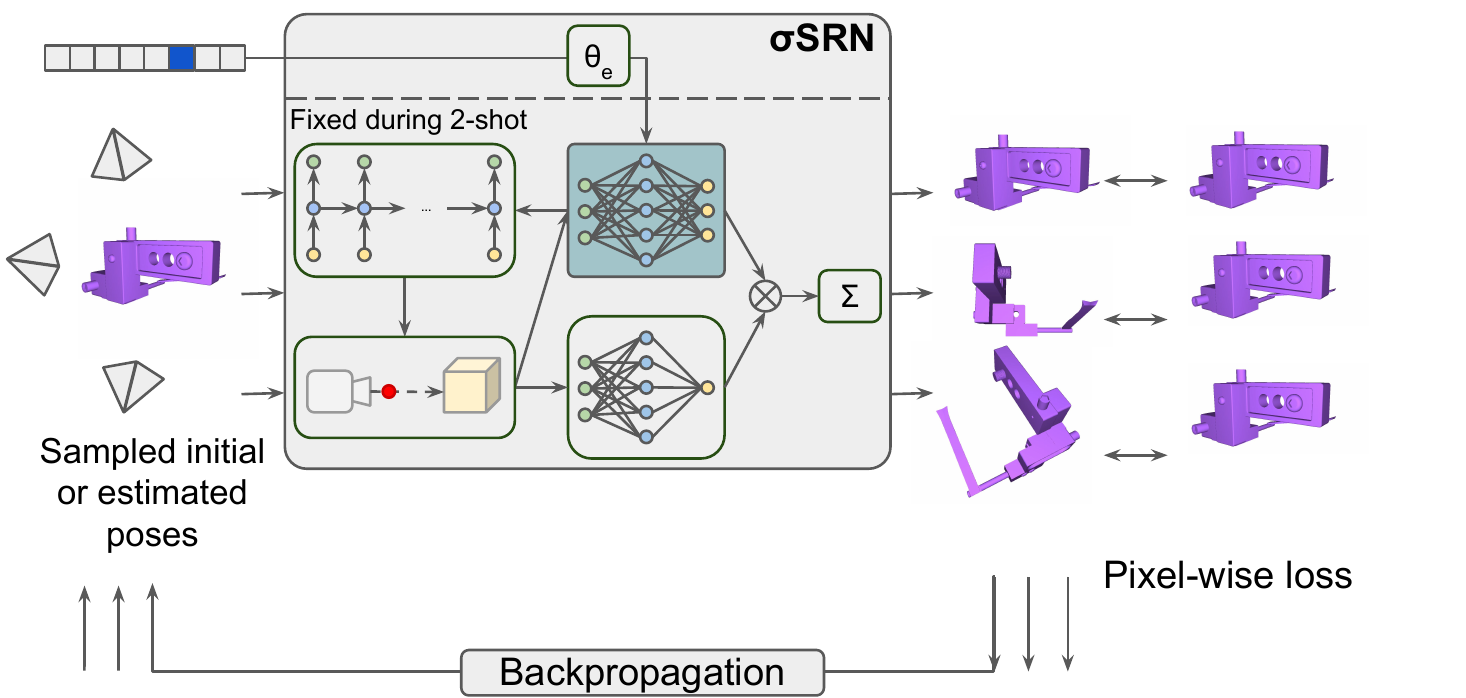}
    }
    \caption{\textbf{Model visualization of i-$\sigma$SRN.} We present an illustration of i-$\sigma$SRN with (a) training a neural renderer, $\sigma$SRN in our case, in phase 1, and (b) estimating the pose by inverting the trained neural renderer in phase 2.}
    \label{fig:model}
    \vspace{-0.5cm}
\end{figure*}
\begin{enumerate}
    \item We present i-$\sigma$SRN, a pose estimation framework that inverts $\sigma$SRN, a novel scene renderer built specifically for pose estimation. $\sigma$SRN extends SRN by incorporating a separate network for density estimation and introduces a new way of obtaining a weighted scene representation over the entire ray trace for each pixel.
    \item We analyze different rendering losses and strategies of initializing pose estimates for pose inference using i-$\sigma$SRN.
    \item We evaluate i-$\sigma$SRN for generalization on objects of seen or unseen categories and compare it with iNerf. We show that our approach outperforms iNerf by a large margin and can generalize to objects of seen or unseen categories.
\end{enumerate}

\section{Related Works}
\label{sec:related}

\paragraph{Implicit Scene Representations and Neural Scene Rendering}
In contrast to explicit scene representations, implicit representations incorporate 3D coordinates as input to a deep neural network enabling resolution-free representation for all topologies.
Neural Radiance Fields (NeRFs) \cite{mildenhall2020nerf} used a model parameterized by the 3D location and viewing direction to predict the color intensities and density at each location in a 3D space.
However limited to one scene per model, this parameterization along with positional encodings allowed NeRFs to generate realistic renders of 3D scenes. Scene representation networks (SRNs)~\cite{sitzmann2019scene} and pixelNeRF~\cite{yu2021pixelnerf} conditioned the occupancy model with a learned low-dimensional embedding representing the scene, which allowed scaling such radiance fields to represent multiple scenes within the same model. While SRN is a promising approach, it uses an autoregressive ray tracer prone to vanishing gradients, making it unsuitable for pose estimation. In this work, we extend SRNs for pose estimation, by incorporating a separate network for density estimation and introduce a new way of obtaining a weighted scene representation over the entire ray trace for each pixel in the render. This shortens the computation path from the input pose to the output render and aids in accurate pose estimation. 

\paragraph{Instance- and Category-Specific Pose Estimation} Most state-of-the-art object pose estimators are either instance-specific~(\cite{huttenlocher1993comparing,hinterstoisser2011gradient,xiang2018posecnn,wang2019densefusion,li2019cdpn,peng2019pvnet,he2020pvn3d}) or category-specific~(\cite{wang2019normalized,ahmadyan2021objectron}). Instance-level pose estimation methods estimate pose parameters of \textit{known} object instances. Early approaches~(\cite{huttenlocher1993comparing,hinterstoisser2011gradient}) require the corresponding CAD models to render templates and match those to learned or hand-crafted features for matching. Learning-based approaches estimate an object's pose by directly regressing the rotation and translation parameters~(\cite{xiang2018posecnn,wang2019densefusion}) and using dense correspondences~(\cite{li2019cdpn}). Keypoint-based approaches~(\cite{peng2019pvnet,he2020pvn3d}) utilized deep neural networks to detect 2D keypoints of an object and computed 6D pose parameters with Perspective-n-Point (PnP) algorithms, improving pose estimates by a large margin. In contrast, recent category-level pose estimation methods~(\cite{wang2019normalized,ahmadyan2021objectron}) estimate poses of unseen object instances within the \textit{known} categories thus addressing generalizability. In contrast to existing deep learning-based pose estimation methods, our approach generalizes to category-level and completely unseen objects.

\paragraph{Generalization methods}
The recently proposed pose estimation methods in \cite{he2022fs6d,liu2022gen6d,sun2022onepose} do not require object CAD models and can predict poses for unseen object categories. In \cite{he2022fs6d}, $k$ support RGB-D images of the same object with a known pose are utilized to estimate the object pose from an RGB-D image, using correspondence feature sets extraction and a point-set registration method. Similarly, \cite{liu2022gen6d} studied the 6-DoF object pose estimation of a novel object using a set of reference poses of the same object and an iterative pose refinement network. OnePose \cite{sun2022onepose} uses a video scan of the novel object to construct the object point cloud using the Structure from Motion (SfM) procedure. Our method does not require any CAD models or depth maps for pose estimation. Also, unlike the ``discriminative'' approaches that filter out information from high dimensional input data, our method takes a ``generative'' approach by utilizing implicit scene representation and preserves scene information to improve generalizability to novel objects.

\section{Background}
\label{sec:background}

\subsection{3D pose representation}
\label{subsec:background pose}
The position and orientation (pose) of a rigid body in 3D space can be defined using six independent variables representing the translation and rotation of rigid body around three independent axes, all with respect to a standard pose. This transformation can then be summarized into a single $4\times4$ matrix that transforms homogeneous coordinates as
\vspace{-0.2cm}
\begin{equation}
    \mathbf{v}^{'} = T\mathbf{v},
\end{equation}
where $\mathbf{v}$ and $\mathbf{v'}$ are the four-dimensional homogeneous coordinates in the original and transformed frame respectively, and $T$ is the $4\times4$ transformation matrix.

There are multiple ways of representing the $3\times3$ rotation sub-matrix of the transformation matrix $T$. We choose to formulate it as the consecutive multiplication of the rotation matrices around the $X, Y, \text{and } Z$ axes (in that order). Appending it with the translation vector we get the transformation
\vspace{-0.2cm}
\begin{equation}
\label{eq:rigid transform}
T = 
\begin{bmatrix}
c_2 c_3 & -c_2 s_3 & s_2 & t_1\\
c_1 s_3 + c_3 s_1 s_2 & c_1 c_3 - s_1 s_2 s_3 & -c_2 s_1 & t_2\\
s_1 s_3 - c_1 c_3 s_2 & c_3 s_1 + c_1 s_2 s_3 & c_1 c_2 & t_3\\
0 & 0 & 0 & 1\\
\end{bmatrix},
\end{equation}
where $c_i$ and $s_i$ represent the cosine and sine of the rotation angles $\theta_i$, $t_i$ represents the translation in three independent directions, and $i$ indexes the $X, Y, \text{and } Z$ axes ($i \in \{1, 2, 3\}$). 
In this paper, the six degrees of freedom that constitute this transformation matrix are inferred from an implicit scene representation model using backpropagated gradients.

\subsection{Scene Representation Networks}
Implicit representations of 3D scenes are parameterized functions that map a point in the 3D space to an occupancy metric. 
These occupancy measures are coupled with a rendering algorithm to generate photo-realistic renders which are regressed towards target RGB images for supervision. In this paper, we build upon the scene representation network (SRN) formulated in \cite{sitzmann2019scene} that uses a hypernetwork approach to generalize to multiple instances of an object. The model uses the camera intrinsics to compute ray directions along which it samples the 3D space using a learned ray marcher, parameterized using an LSTM~\cite{hochreiter1997long} module. The end-point of this ray is fed into a scene representation network whose parameters are further parameterized using a hypernetwork conditioned on the unique index of the object in the training set. The representation is then fed into a pixel generation network that outputs the RGB values one pixel at a time. 

\subsection{Pose Estimation by Inverting a Neural Renderer}
Given a query image and a scene representation model that can render images at arbitrary camera poses, we are interested in inferring the camera pose utilized in obtaining the query image. Despite its obvious potential benefits in out-of-distribution generalization, this formulation has only recently become a topic of interest in deep learning literature. Yen et al.~\cite{yen2021inerf} proposed iNeRF that inverts a pixelNeRF \cite{yu2021pixelnerf} to obtain the pose estimate from an RGB image. They formalized the problem of obtaining the camera pose as
\vspace{-0.2cm}
\begin{equation}
\label{eq:inerf formulation}
    T^* = \argmin_{T \in \text{SE(3)}}\ L(\mathbf{im}_{\text{pred}}, \mathbf{im}_{\text{input}}),
\end{equation}
where SE(3) denotes the group of all rigid transformations in 3D, $\mathbf{im}_{\text{pred}}$ and $\mathbf{im}_{\text{input}}$ denote the output render and query image respectively, and $L(\cdot)$ is a loss function that drives the only source of supervision towards the pose estimate.

The way Yu et al.~\cite{yu2021pixelnerf} addressed the problem uncovered many open challenges that need to be addressed for this approach to scale. They argued that carefully sampled rays within an interest region are critical for this approach to work. However, we show that with our parameterization of the input pose there is no need for such sampling, and that losses on the entire image can provide sufficient supervision for pose estimation. Additionally, iNeRF demonstrated experiments assuming an initial pose available within $30^{\circ}$ of the target pose, whereas we explore strategies that work around any such assumptions.
\section{Methodology}
\label{sec:method}

\begin{figure}[t]
    \centering
    \subcaptionbox{Fixed initial estimates of the camera pose, uniformly spanning three latitudnal lines on a sphere.}
    {
        \includegraphics[width=0.4\linewidth]{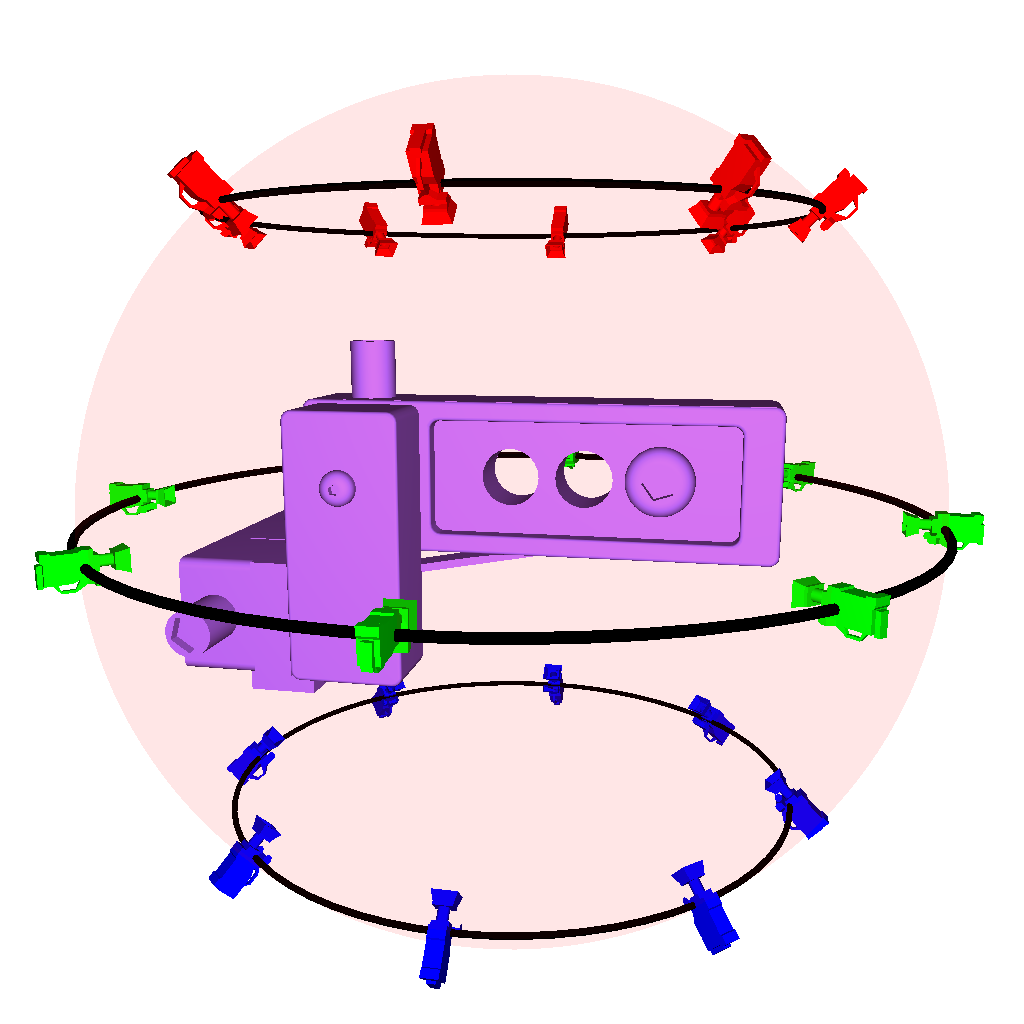}
    }
    \hfill
    \subcaptionbox{Initial estimates of the camera pose in the neighbourhood of the target pose.}
    {
        \includegraphics[width=0.4\linewidth]{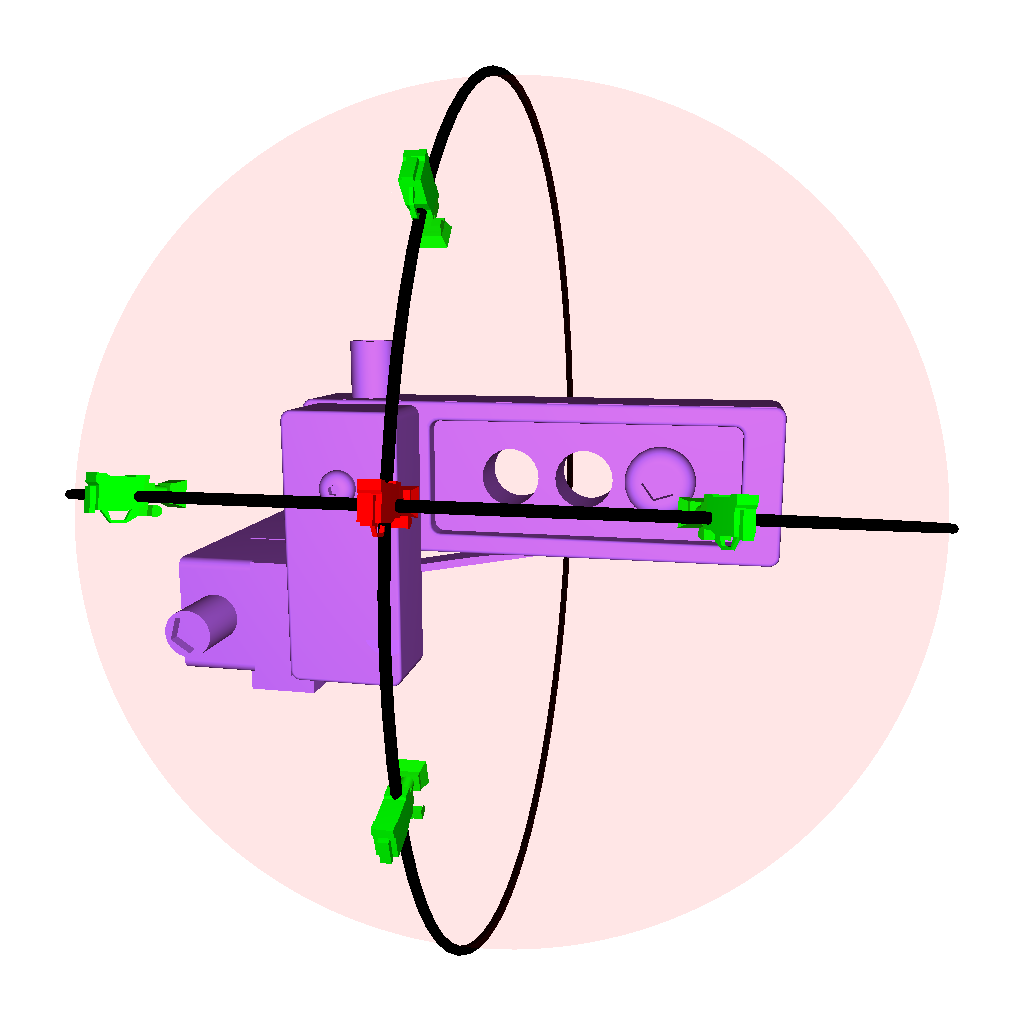}
    }
    \caption{\textbf{Initial estimates of the camera pose.} We investigate two different ways of sampling initial poses for inference with i-$\sigma$SRN.}
    \label{fig:initial_pose_sampling}
    \vspace{-0.5cm}
\end{figure}
\begin{figure*}[t]
     \centering
     \begin{subfigure}[b]{0.44\linewidth}
         \centering
         \includegraphics[width=0.45\linewidth]{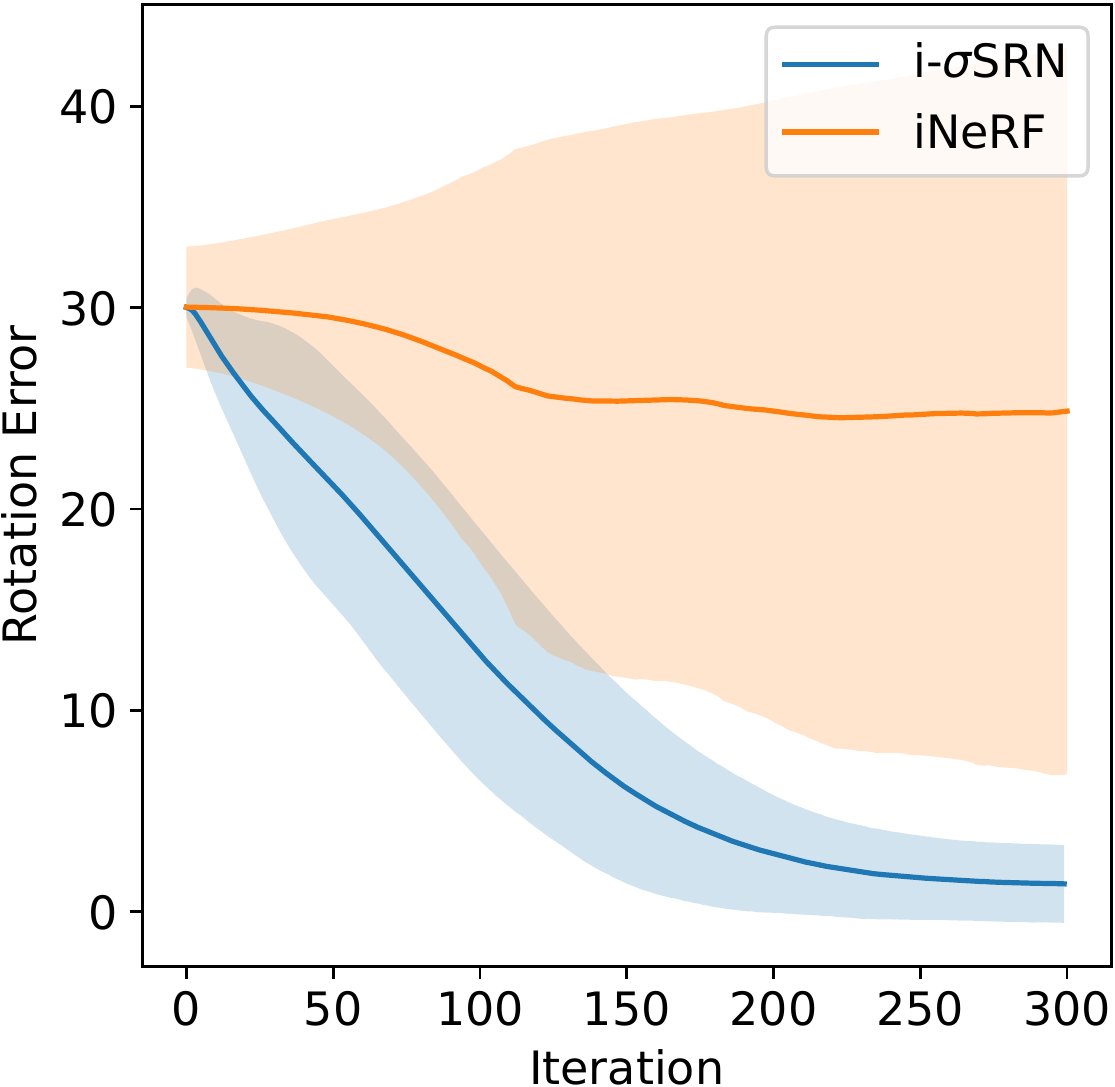}
     \hfill
         \centering
         \includegraphics[width=0.44\linewidth]{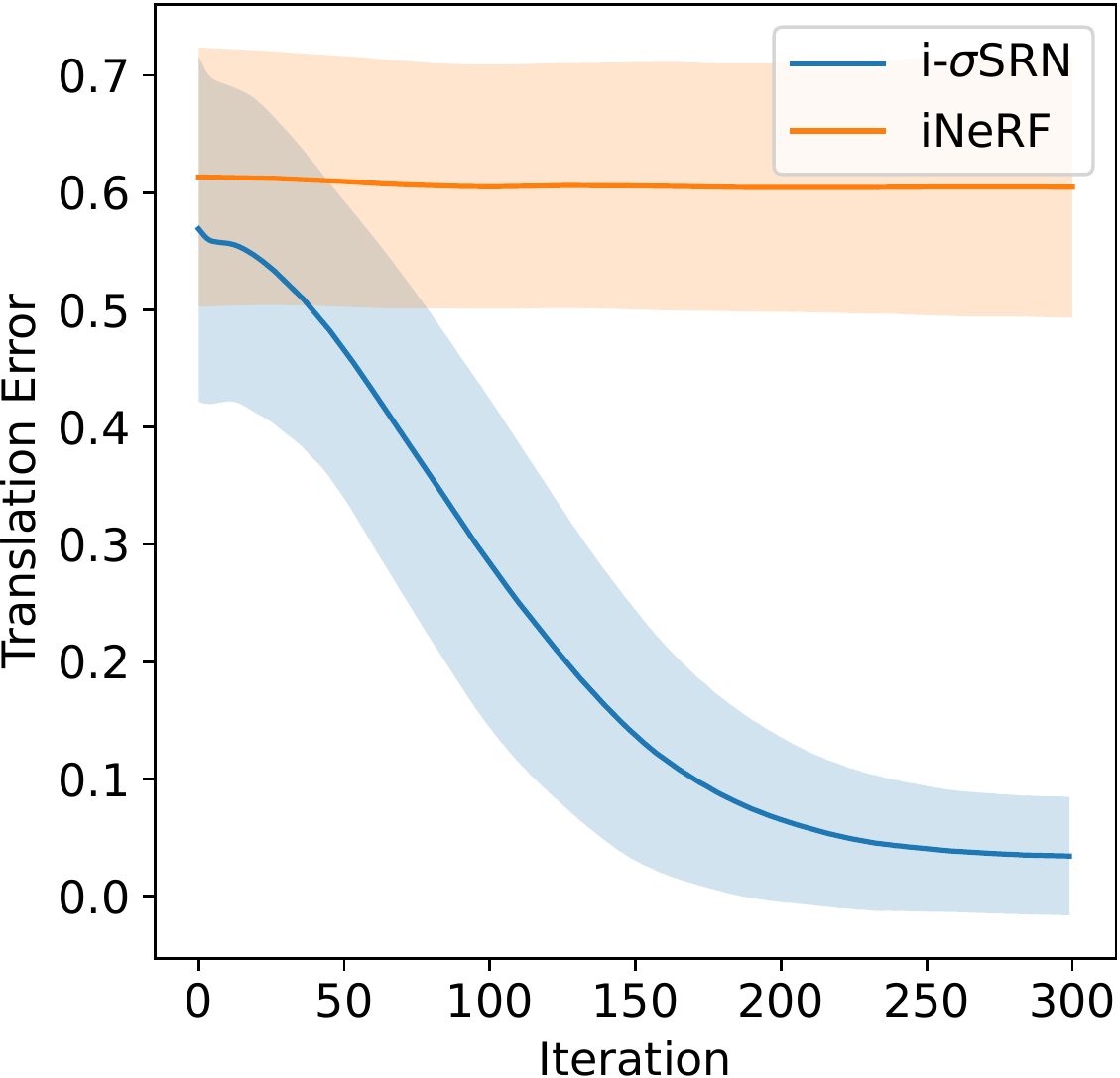}
         \caption{Cars.}
         \label{subfig:car_plot}
     \end{subfigure}
     \begin{subfigure}[b]{0.44\linewidth}
         \centering
         \includegraphics[width=0.45\linewidth]{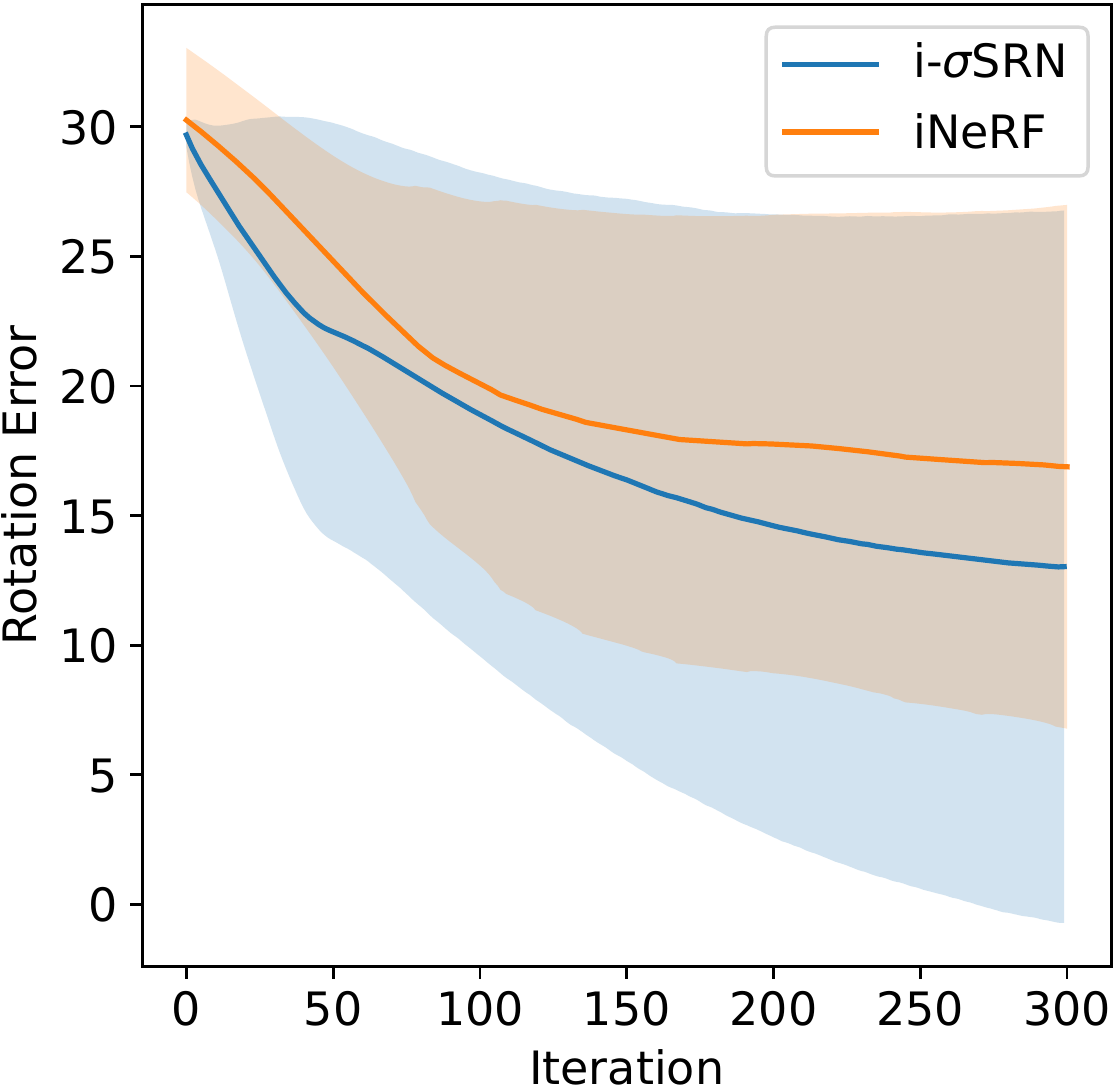}
     \hfill
         \centering
         \includegraphics[width=0.44\linewidth]{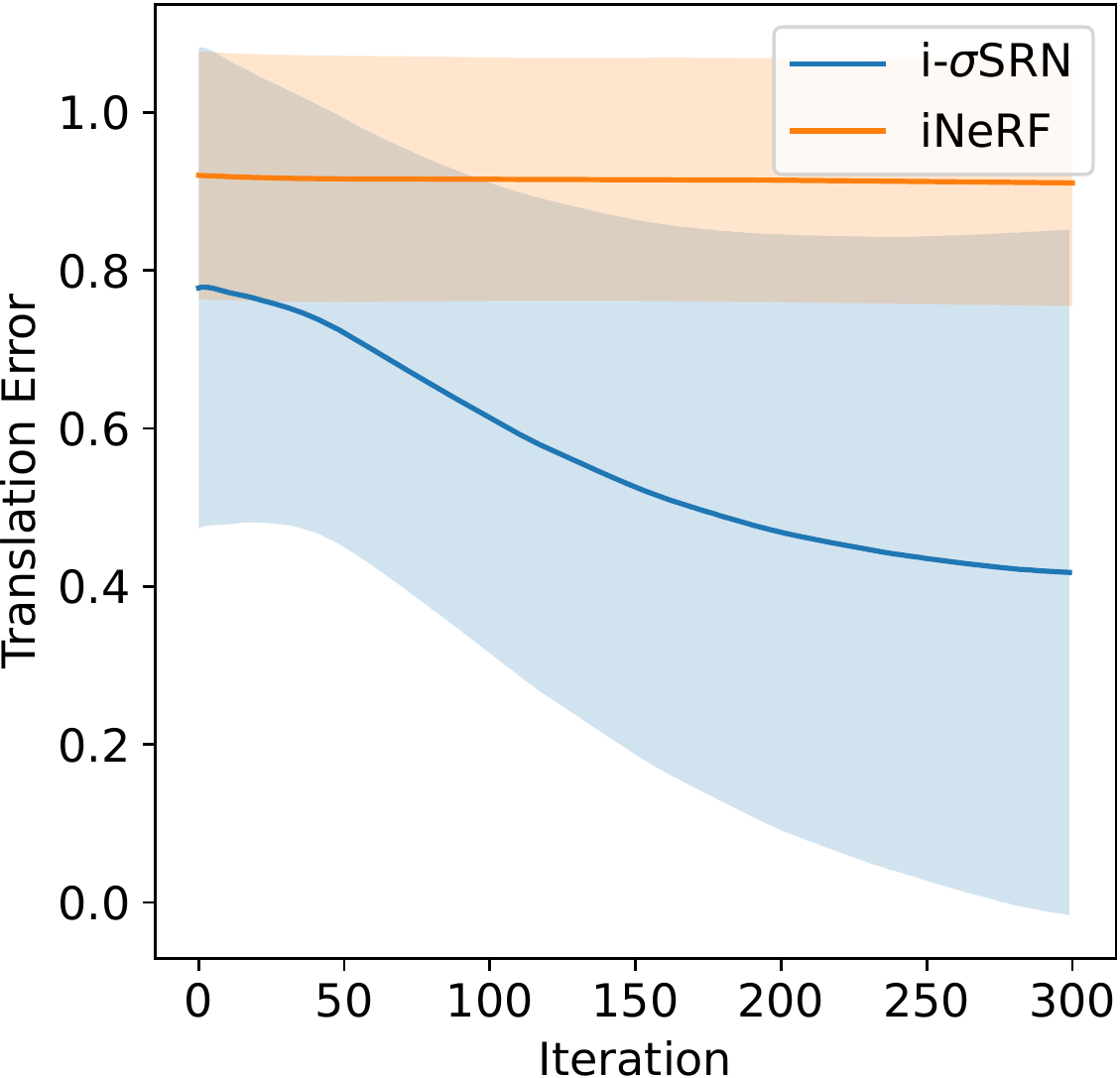}
         \caption{Chairs.}
         \label{subfig:chair_plot}
     \end{subfigure}
    \caption{\textbf{Comparing rotation and translation refinement on ShapeNet cars and chairs.} We illustrate the mean pose errors with 1 standard deviation (shaded) as evaluation progresses.}
    \label{fig:car_chair_plots}
    \vspace{-0.5cm}
\end{figure*}

Inferring the object pose from an RGB image is central to many robotic applications. Most methods for pose estimation use a ``discriminative'' approach where they learn a feed-forward model that filters out information from high-dimensional inputs (such as images, point clouds, depth maps) to predict the pose. However, these approaches suffer from generalization issues during test time. We take a ``generative'' approach to pose estimation that retains maximal scene information within the model, helping with better generalization. We build a scene representation model that can generate the object view from any query pose, and during test time, the model uses its rendering capability to infer the pose of a query object image by backpropagating through the scene generation model.

\subsection{Learning Scene Representations}

Building upon the scene representation networks (SRNs) in \cite{sitzmann2019scene}, we propose $\sigma$SRN, a scene representation model that can render scenes at unknown poses but with a shorter gradient path for subsequent support for pose estimation. We illustrate our model in Figure \ref{fig:model}a.

The scene representation model takes camera intrinsics K and a query extrinsic pose $\{\theta_i, t_i\}_{i=1}^3$ as input. The camera intrinsics help the model derive direction vectors along which to trace the ray corresponding to each pixel on the focal plane. The ray trace is initialized with coordinates $(x^{(0)},y^{(0)},z^{(0)})$ randomly distributed close to the focal plane of the camera. We feed these 3D coordinates into a scene representation model that generates a vector representing the scene at these coordinates.
To enable storing information about multiple object instances in the same model, parameters of the scene representation model are further parameterized using a hypernetwork that is conditioned on a unique index that identifies that instance among all instances in the training data. This computation can be summarized as,
\vspace{-0.2cm}
\begin{equation}
    \phi^{(i)} = f\bigg( (x^{(i)},y^{(i)},z^{(i)});\ \theta_{f}^{\text{hyp}}(\theta_e\cdot\text{onehot}(\iota); \theta_{f}) \bigg),
\end{equation}
where $(x^{(i)},y^{(i)},z^{(i)})$ are the i-th coordinates along the traced ray, $f(\cdot)$ is a learnable function modeled using a neural network with parameters output from a hypernetwork $\theta_{f}^{\text{hyp}}(\cdot)$. Input to the hypernetwork is an embedding vector obtained by multiplying a parameter matrix $\theta_e$ with a one-hot vector that has a single high at index $\iota$, essentially returning the $\iota$-th column of $\theta_e$. $\phi^{(i)}$ is then fed into a LSTM~\cite{hochreiter1997long} module $r(\cdot)$ that generates the next coordinates on the ray trace,
\begin{equation}
    (x^{(i+1)},y^{(i+1)},z^{(i+1)}) = (x^{(i)},y^{(i)},z^{(i)}) + r(\phi^{(i)}; \theta_r).
\end{equation}
In addition to the scene representation module, we introduce a density prediction network that predicts the density of space at the generated 3D coordinates in the ray trace. We define this density $\sigma$ as
\begin{equation}
    \sigma^{(i)} = g((x^{(i)},y^{(i)},z^{(i)}); \theta_{g}),
\end{equation}
where $g(\cdot)$ is learnable function modeled using a neural network with parameters $\theta_{g}$. After obtaining both the description of the 3D point and the predicted density of $M$ discrete samples along the ray trace, we multiply them together to obtain the final representation vector for the pixel, given as
\vspace{-0.2cm}
\begin{equation}
    \phi = \sum_{i=1}^{M} \sigma^{(i)} * \phi^{(i)}.
\end{equation}
Note here that $\sigma^{(i)}$ are scalars and $\phi^{(i)}$ are multi-dimensional vectors, and $*$ represents scalar-vector multiplication.
Since the final scene representation $\phi$ is a function of the entire trace and not just the end-point, this shortens the computation path from the input pose to the output rendering, preventing the gradient from vanishing during subsequent pose estimation using backpropagation.

Finally, we use a convolutional pixel generator to obtain the RGB output at each pixel in the output image,
\begin{equation}
    \mathbf{im}_{\text{pred}} = h(\phi'; \theta_h),
\end{equation}
where $h$ is modeled using a neural network with parameters $\theta_h$. We train this scene representation model end-to-end using the mean-squared error, along with latent regularization, between the input query image and the output render to obtain the optimal parameter set $(\theta_e, \theta_{f}, \theta_{g}, \theta_{r}, \theta_{h})$.

\begin{table*}[htbp]
\caption{Pose estimation errors on the Shapenet Cars, ShapeNet Chairs and AssemblyPose dataset.}
\centering
\resizebox{0.73\linewidth}{!}{
\begin{tabular}{llcccccccc}
\toprule
\textbf{Method} & \textbf{Inference} & \multicolumn{2}{c}{\textbf{Cars}} & & \multicolumn{2}{c}{\textbf{Chairs}} & & \multicolumn{2}{c}{\textbf{AssemblyPose}}\\
\cmidrule{3-4} \cmidrule{6-7}  \cmidrule{9-10} 
 & &  \textbf{\textit{Rotation}} & \textbf{\textit{Translation}} & & \textbf{\textit{Rotation}} & \textbf{\textit{Translation}} & & \textbf{\textit{Rotation}} & \textbf{\textit{Translation}} \\
\midrule
\textbf{iNeRF} & single & 32.23 & 0.61 & & 17.30 & 0.93 & & 46.66 & 0.55 \\
\textbf{iNeRF} & multiple (10) & 24.88 & 0.60 & & 16.87 & 0.91 & & 46.13 & 0.52 \\
\textbf{i-$\sigma$SRN (ours)} & multiple (fixed, 24) & 2.60 & 0.06 & & 30.54 & 0.79 & & 3.59 & 0.16 \\
\textbf{i-$\sigma$SRN (ours)} & multiple (neighbor, 4) & \textbf{1.38} & \textbf{0.03} & & \textbf{12.87} & \textbf{0.40} & & \textbf{2.36} & \textbf{0.04}\\
\bottomrule
\end{tabular}
}
\label{tab:pose-all}
\end{table*}
\begin{table*}[t]
\caption{Two-shot pose estimation errors on the ShapeNet Cars, ShapeNet Chairs and AssemblyPose dataset.}
\vspace{-0.1cm}
\centering
\resizebox{0.73\linewidth}{!}{
\begin{tabular}{llcccccccc}
\toprule
\textbf{Method} & \textbf{Inference} & \multicolumn{2}{c}{\textbf{Cars}} & 
 & \multicolumn{2}{c}{\textbf{Chairs}} & & \multicolumn{2}{c}{\textbf{AssemblyPose}}\\
\cmidrule{3-4} \cmidrule{6-7} \cmidrule{9-10}
 & & \textbf{\textit{Rotation}} & \textbf{\textit{Translation}} & & \textbf{\textit{Rotation}} & \textbf{\textit{Translation}} & & \textbf{\textit{Rotation}} & \textbf{\textit{Translation}} \\
\midrule
\textbf{iNeRF} & 2-shot samples & 105.78 & 1.47 & & 79.31 & 2.03 & & 83.76 & 1.09 \\
\textbf{i-$\sigma$SRN (ours)} & multiple (fixed, 24) & 38.37 & 0.66 & & 33.52 & 0.91 & & 63.69 & 0.79\\
\textbf{i-$\sigma$SRN (ours)} & multiple (neighbor, 4) & \textbf{14.59} & \textbf{0.34} & & \textbf{13.03} & \textbf{0.42} & & \textbf{32.48} & \textbf{0.48} \\
\bottomrule
\end{tabular}
}
\vspace{-0.5cm}
\label{tab:few-shot}
\end{table*}

\subsection{Pose Estimation}
\label{subsec:method pose estimation}

Here we address the problem of estimating the pose of the camera that was used to capture the image of an object, while given a learned implicit representation of the scene equipped with a differentiable renderer. We present i-$\sigma$SRN, a pose estimation algorithm that is formulated as
\vspace{-0.1cm}
\begin{equation}
    \label{eq:6dof estimation formulation}
    \argmin_{\{\theta_i, t_i\}_{i=1}^3}\ L(\mathbf{im}_{\text{pred}}, \mathbf{im}_{\text{input}}).
\end{equation}
We investigate different loss functions $L(\cdot)$ during evaluation. Unlike the formulation in iNeRF that optimizes for all 16 entries in the rigid transformation matrix (see Eq.~\ref{eq:inerf formulation}), our formulation optimizes for just the 6 DoF that a rigid object can rotate/translate in. This allows for an easier optimization since the search space is much smaller and the renderer is constrained to render only rigid transformations of the object.

After training the implicit scene representation model, we freeze the entire model and optimize for the 6 DoF pose $\{\theta_i, t_i\}_{i=1}^3$. Our model starts with an initial guess of the pose and generates a render. It then optimizes the loss between the output render and the query image w.r.t. the six extrinsic parameters using gradient descent updates from the Adam \cite{kingma2014adam} optimizer. We parallelize this optimization over a batch of query images using per-sample gradients. We illustrate i-$\sigma$SRN in Figure \ref{fig:model}b. We use two evaluation protocols to estimate the camera pose. We take 24 initial guesses of the pose as shown in Figure \ref{fig:initial_pose_sampling}a. Centered around the object, this is eight equally spaced camera poses along the 45\textdegree\ latitude, the equator, and -45\textdegree\ latitude. For each pose, the camera points to the object's center. We use backpropagation and 300 iterations from each initial pose, then take the solution with the lowest loss as the estimated pose.
The second approach uses four initial guesses of the camera pose as shown in Figure \ref{fig:initial_pose_sampling}b. Here we assume we’re given a rough estimate of the pose that lets us create four initial poses offset by 30\textdegree, above, below, left, and right of the estimate, respectively. We use backpropagation and 300 iterations from each initial pose, then take the solution with the lowest loss as the estimated pose.

\subsection{Two-shot Generalization}

Training the $\sigma$SRN (with parameters $(\theta_e, \theta_f, \theta_g, \theta_{r}, \theta_h)$) on a diverse set of instances helps the model learn unique embeddings corresponding to each instance in the dataset. The problem that we target here is that of pose estimation of unseen object instances during test time.
In order to generalize to new instances, we fine-tune the $\sigma$SRN, in the same fashion as a vanilla SRN, by freezing all parameters except the embedding vectors that correspond to the unique indices of instances in the dataset. That is, we solve the minimization problem,
\vspace{-0.2cm}
\begin{equation}
    \hat\theta_e = \argmin_{\theta_e} \ ||\mathbf{im}_{\text{pred}} - \mathbf{im}_{\text{input}}||_2^2.
\end{equation}

After having obtained a new embedding corresponding to the novel instances, we evaluate pose estimation using the method described in Section \ref{subsec:method pose estimation} with parameter set $(\hat\theta_e, \theta_f, \theta_g, \theta_{r}, \theta_h)$.

\section{Evaluation}
\label{sec:experiments}

\subsection{Experimental Settings}

\paragraph{Datasets}
We demonstrate i-$\sigma$SRN on the ShapeNetv2 cars and chairs datasets \cite{chang2015shapenet} made available by \cite{sitzmann2019scene} and a collection of CAD shapes taken from the Autodesk Fusion 360 Gallery dataset \cite{willis2022joinable} that we call ``AssemblyPose'' in this context. We use the training dataset for category-specific and 2-shot training of the neural renderer, whereas for inference, we apply different splits (category-specific, 2-shot). There are 2151 cars, 4612 chairs, and 107 AssemblyPose shapes in the training set. Each shape is scaled to unit length along the diagonal of its bounding box. The observations are rendered images of the shapes from the camera randomly placed on a sphere centered on the object. The camera pose is pointing towards the object center with no roll. The sphere radius is 1.3, 2, and 1 for the cars, chairs, and assemblies respectively. The testing dataset has the same shapes and the rendered observations are from camera positions along a sampling of the spherical spiral. The camera points towards the object center for each sampled position with no roll. The spiral radius is the same as the training set. For the category-specific experiments, we report the average errors on ten models and ten unknown poses per model from the validation set (not used during training). For our 2-shot generalization experiments, we re-train the embedding parameters of the $\sigma$SRN model on 2 observations from a set of new shapes. The number of cars, chairs and assemblies in this new collection are 352, 362 and 30, respectively.

\paragraph{Pose estimation comparison}
The work most comparable to ours is iNeRF~\cite{yen2021inerf} that we will use for comparison. Their work used pixelNeRF as the neural renderer for pose estimation. We used the provided pre-trained weights for ShapeNet Cars and Chairs and followed similar training settings for AssemblyPose. For category-specific evaluation, we uniformly sampled the source renderings (single=1, multiple=10) from the train dataset for inference. For 2-shot evaluation, we used the 2-shot sample as source renderings for inference. We choose the best pose estimation based on the lowest loss.

\paragraph{Evaluation metrics}
We report the rotation and translation error between the predicted and the target camera pose,
\vspace{-1.5em}
\begin{align}
    e_{\text{tra}} &= || \mathbf{t_{gt}} - \mathbf{t_{pred}} ||_2 \textrm{, and}\\
    e_{\text{rot}} &= \cos^{-1} \bigg( 0.5 * (\text{tr}( \mathbf{R_{pred}} \mathbf{R_{gt}^{-1}} ) - 1) \bigg).
\end{align}
Rotation errors are reported in degrees. With the shapes scaled to unit size, the translation errors can be interpreted as a percentage of the shape size. 

\paragraph{Loss functions}
We test \textit{mean absolute error} (MAE), \textit{mean squared error} (MSE) and \textit{gradient magnitude similarity deviation} \cite{gmsd} loss (GMSD) as the loss function in Eq. \ref{eq:6dof estimation formulation}, with the 24 fixed-based estimation protocol described in Section \ref{subsec:method pose estimation}. We choose \textit{mean absolute error} as the loss function for our experiments based on its top score relative to the other loss functions (see Table \ref{tab:eval_losses}).

\paragraph{Training and Optimization}
We train separate $\sigma$SRN models for the cars, chairs, and assemblies. Fifty observations per car and chair instance are used to train each model for 8 and 9 epochs, respectively. One thousand five hundred observations per assembly instance are used to train its model to 12 epochs. A batch size of 10 on an NVIDIA V100 GPU with PyTorch, a fixed learning rate of 5e-5, and Adam \cite{kingma2014adam} optimizer is used for training. The input and output image resolution is 64$\times$64. For 2-shot training, we use the same $\sigma$SRN training setup. With the smaller dataset and only the instance embeddings to train, we train to approximately 20\% of the steps used for $\sigma$SRN.
For i-$\sigma$SRN, we use the Adam optimizer with a fixed learning rate of 1e-1. Evaluations are run in a batch size of 5 on an NVIDIA V100 GPU. For each pose estimation, we run both i-$\sigma$SRN and iNeRF for up to 300 optimization steps. In our experiments, an optimization step of i-$\sigma$SRN took on average \textbf{115 ms} (batch size 5), while iNeRF required 323 ms (batch size 1).

\subsection{Main Results \& Analysis}

\paragraph{Pose Estimation Performance}
We present results for the cars, chairs, and assemblies pose estimation tests and a comparison of the i-$\sigma$SRN to the iNeRF errors in Table \ref{tab:pose-all}. Our 4-neighbor testing protocol produced the best results for all three data sets. We compare the rotation and translation refinement trajectories of i-$\sigma$SRN with iNeRF in Figure \ref{fig:car_chair_plots}, in which we clearly see that i-$\sigma$SRN converges to far more accurate poses that iNeRF on both ShapeNet datasets.

\begin{table}[t]
\caption{Average rotation error associated with MAE, MSE and GMSD losses with 24 fixed initial camera poses.}
\vspace{-0.1cm}
\begin{center}
\begin{tabular}{lccc}
\toprule
\textbf{Dataset}&\textbf{MAE}&\textbf{MSE}&\textbf{GMSD} \\
\midrule
\textbf{Cars} & \textbf{2.60} & 2.76 & 9.79 \\
\textbf{Chairs} & 30.54 & 27.82 & \textbf{16.68} \\
\textbf{AssemblyPose} & \textbf{3.59} & 4.15 & 6.19 \\
\bottomrule
\end{tabular}
\label{tab:eval_losses}
\vspace{-0.5cm}
\end{center}
\end{table}

\begin{figure}[t]
     \centering
     \begin{subfigure}[b]{0.2\linewidth}
         \centering
         \includegraphics[width=\linewidth]{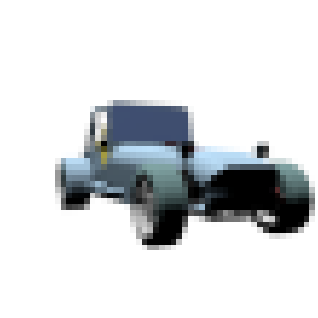}
         \renewcommand{\thesubfigure}{a}%
         \caption{Target}
         \label{fig:conv_B_2shot_a}
     \end{subfigure}
     \hfill
     \begin{subfigure}[b]{0.2\linewidth}
         \centering
         \includegraphics[width=\linewidth]{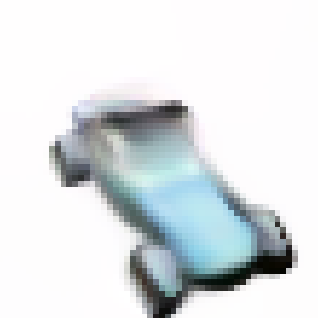}
         \renewcommand{\thesubfigure}{b}%
         \caption{Init. pose}
         \label{fig:conv_B_2shot_b}
     \end{subfigure}
     \hfill
     \begin{subfigure}[b]{0.2\linewidth}
         \centering
         \includegraphics[width=\linewidth]{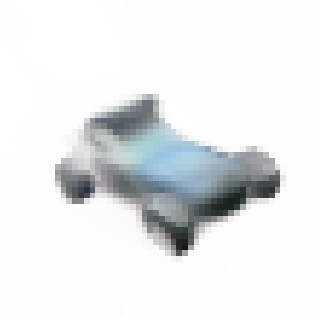}
         \renewcommand{\thesubfigure}{c}%
         \caption{100 steps}
         \label{fig:conv_B_2shot_c}
     \end{subfigure}
     \hfill
     \begin{subfigure}[b]{0.2\linewidth}
         \centering
         \includegraphics[width=\linewidth]{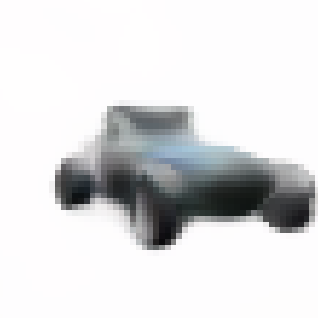}
         \renewcommand{\thesubfigure}{d}%
         \caption{300 steps}
         \label{fig:conv_B_2shot_d}
     \end{subfigure}
    \caption{\textbf{Two-shot generalization for pose estimation.} (a) is the target image of an unseen object, and (b)-(d) are snapshots to the pose refinement trajectory with lowest MAE.}
    \label{fig:convergence_2shot}
    \vspace{-0.5cm}
\end{figure}

\paragraph{Two-shot Pose Estimation Performance}
The 2-shot generalization test results are presented in Table \ref{tab:few-shot}. Figure \ref{fig:convergence_2shot} shows an example of camera pose convergence from the 2-shot cars experiment. Our 4-neighbor testing protocol produced the best results for all three data sets.  

\begin{figure}[t]
     \centering
     \begin{subfigure}[b]{0.2\linewidth}
         \centering
         \includegraphics[width=\linewidth]{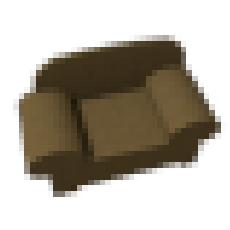}
         \caption{Target}
         \label{fig:color_a}
     \end{subfigure}
     \hfill
     \begin{subfigure}[b]{0.2\linewidth}
         \centering
         \includegraphics[width=\linewidth]{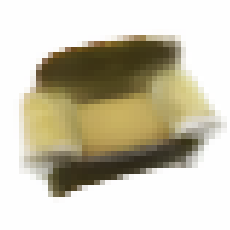}
         \caption{$\uparrow$ MAE}
         \label{fig:color_b}
     \end{subfigure}
     \hfill
     \begin{subfigure}[b]{0.2\linewidth}
         \centering
         \includegraphics[width=\linewidth]{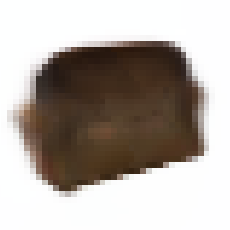}
         \caption{$\downarrow$ MAE}
         \label{fig:color_c}
     \end{subfigure}
     \hfill
     \begin{subfigure}[b]{0.2\linewidth}
         \centering
         \includegraphics[width=\linewidth]{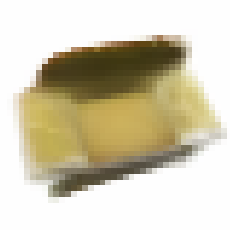}
         \caption{GMSD}
         \label{fig:colr_d}
     \end{subfigure}
    \caption{\textbf{The effect of color on loss.} (a) is the target image, (b) is one of the 24 solutions with the best pose estimate, (c) is the solution with the lowest MAE loss, (and 180\degree rotation error), and (d) is the solution with the best GMSD loss.}
    \label{fig:color_challenge}
    \vspace{-0.5cm}
\end{figure}

\paragraph{Effect of loss function choice on performance}
The loss function guides the optimization to find the target camera pose. We run multiple optimizations from different starting poses and rely on the lowest loss to indicate the solution for the pose estimate. When the rendered image matches closely with the target image in color and shape, a simple MAE loss works well for both criteria. The car data set is an example of where this is the case. The chairs data set, however, has several instances where $\sigma$SRN struggles to render its proper color. This leads to solutions that converge to the incorrect pose but have the lowest loss. For example, Figure \ref{fig:color_c} shows the $\sigma$SRN rendered image from the camera pose associated with the lowest MAE loss. This has a rotation error of approximately 180\textdegree. A solution does converge on the target (see Figure \ref{fig:color_b}) but its MAE loss is high because of the incorrect coloring in the rendering. We test the GMSD loss to see if image gradient information is less sensitive to coloring errors. It is better than MAE loss for cases like the discolored chair rendering, but in general it did not perform as well as MAE loss. The challenge is finding a loss function that can account for both situations. 

\begin{figure}[t]
     \centering
     \begin{subfigure}[b]{0.2\linewidth}
         \centering
         \includegraphics[width=\linewidth]{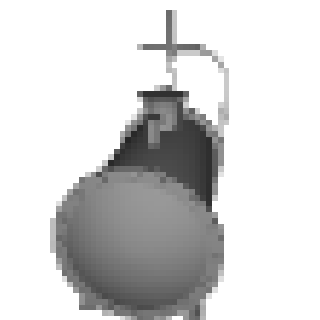}
         \caption{Target}
         \label{fig:2shot_a}
     \end{subfigure}
     \hfill
     \begin{subfigure}[b]{0.2\linewidth}
         \centering
         \includegraphics[width=\linewidth]{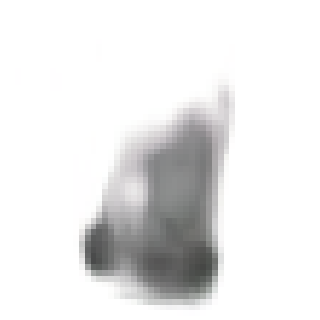}
         \caption{Init. pose}
         \label{fig:2shot_b}
     \end{subfigure}
     \hfill
     \begin{subfigure}[b]{0.2\linewidth}
         \centering
         \includegraphics[width=\linewidth]{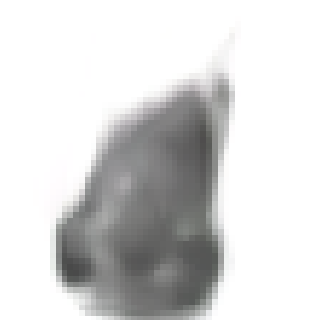}
         \caption{100 steps}
         \label{fig:2shot_c}
     \end{subfigure}
     \hfill
     \begin{subfigure}[b]{0.2\linewidth}
         \centering
         \includegraphics[width=\linewidth]{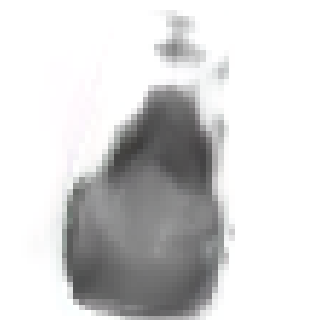}
         \caption{300 steps}
         \label{fig:2shot_d}
     \end{subfigure}
    \caption{Degraded render quality with 2-shot AssemblyPose.}
    \label{fig:2shot_challenge}
    \vspace{-0.5cm}
\end{figure}

\paragraph{Effect of initial pose sampling on performance} 
The 4-neighbor testing protocol yields better results than the 24-fixed protocol. We believe this is because $\sigma$SRN renderings appear similar in the same camera pose neighborhood. This enables the best pose estimate to have a lower loss than its neighbors. We also find that the rendering quality at the initial pose is not crucial.  This is illustrated in Figure \ref{fig:2shot_challenge} from the AssemblyPose data set. The shapes vary drastically and make for limited generalization for 2-shot performance. Despite the low-quality rendering at the starting pose, the gradient still moves the camera to the target, where we see the quality of the rendering improve. 

\section{Conclusion}
\label{sec:conclusion}
We proposed a novel method for estimating the pose of an object given an RGB image, an implicit representation of the 3D scene, and a differentiable neural scene renderer.
Building upon SRNs~\cite{sitzmann2019scene} we proposed $\sigma$SRN, an impicit scene representation model with a shorter computation path between the input pose and the output render, suitable for pose estimation by model inversion. We presented i-$\sigma$SRN, a pose estimation method that inverts the $\sigma$SRN to obtain accurate pose estimates.
We evaluated our work on three different datasets under different experimental settings with iNeRF~\cite{yen2021inerf}, and showed that our model is significantly faster and converges to a pose with lower error on all datasets, even without any initial pose assumptions, as made by iNeRF. We provided an initialization strategy for pose estimation and analyze how different image losses affect pose estimation performance. We also showcased the generalization capability of our approach, where our model was able to estimate the camera pose of a novel instance given only two observations for fine-tuning. Future work includes testing the system with real cameras and robots for real-time pose estimation in the clutter with occluded objects. This adaptation to a real-world setup could be achieved by scaling the neural architectures and training on multi-view captures. Additionally, it would also be interesting to test depth maps as another rendering mode to use with depth cameras.

\addtolength{\textheight}{0cm}   



\clearpage
\clearpage
\bibliographystyle{IEEEtran}
\bibliography{references}

\begin{thebibliography}{10}
\providecommand{\url}[1]{#1}
\csname url@rmstyle\endcsname
\providecommand{\newblock}{\relax}
\providecommand{\bibinfo}[2]{#2}
\providecommand\BIBentrySTDinterwordspacing{\spaceskip=0pt\relax}
\providecommand\BIBentryALTinterwordstretchfactor{4}
\providecommand\BIBentryALTinterwordspacing{\spaceskip=\fontdimen2\font plus
\BIBentryALTinterwordstretchfactor\fontdimen3\font minus
  \fontdimen4\font\relax}
\providecommand\BIBforeignlanguage[2]{{%
\expandafter\ifx\csname l@#1\endcsname\relax
\typeout{** WARNING: IEEEtran.bst: No hyphenation pattern has been}%
\typeout{** loaded for the language `#1'. Using the pattern for}%
\typeout{** the default language instead.}%
\else
\language=\csname l@#1\endcsname
\fi
#2}}

\bibitem{litvak2019learning}
Y.~Litvak, A.~Biess, and A.~Bar-Hillel, ``Learning pose estimation for
  high-precision robotic assembly using simulated depth images,'' in \emph{2019
  International Conference on Robotics and Automation (ICRA)}.\hskip 1em plus
  0.5em minus 0.4em\relax IEEE, 2019, pp. 3521--3527.

\bibitem{chen2020repetitive}
C.~Chen, T.~Wang, D.~Li, and J.~Hong, ``Repetitive assembly action recognition
  based on object detection and pose estimation,'' \emph{Journal of
  Manufacturing Systems}, vol.~55, pp. 325--333, 2020.

\bibitem{rocha2014object}
L.~F. Rocha, M.~Ferreira, V.~Santos, and A.~P. Moreira, ``Object recognition
  and pose estimation for industrial applications: A cascade system,''
  \emph{Robotics and Computer-Integrated Manufacturing}, vol.~30, no.~6, pp.
  605--621, 2014.

\bibitem{choi2012voting}
C.~Choi, Y.~Taguchi, O.~Tuzel, M.-Y. Liu, and S.~Ramalingam, ``Voting-based
  pose estimation for robotic assembly using a 3d sensor,'' in \emph{2012 IEEE
  International Conference on Robotics and Automation}.\hskip 1em plus 0.5em
  minus 0.4em\relax IEEE, 2012, pp. 1724--1731.

\bibitem{chen2018patient}
K.~Chen, P.~Gabriel, A.~Alasfour, C.~Gong, W.~K. Doyle, O.~Devinsky,
  D.~Friedman, P.~Dugan, L.~Melloni, T.~Thesen, \emph{et~al.},
  ``Patient-specific pose estimation in clinical environments,'' \emph{IEEE
  journal of translational engineering in health and medicine}, vol.~6, pp.
  1--11, 2018.

\bibitem{obdrvzalek2012accuracy}
{\v{S}}.~Obdr{\v{z}}{\'a}lek, G.~Kurillo, F.~Ofli, R.~Bajcsy, E.~Seto,
  H.~Jimison, and M.~Pavel, ``Accuracy and robustness of kinect pose estimation
  in the context of coaching of elderly population,'' in \emph{2012 Annual
  International Conference of the IEEE Engineering in Medicine and Biology
  Society}.\hskip 1em plus 0.5em minus 0.4em\relax IEEE, 2012, pp. 1188--1193.

\bibitem{tremblay2018deep}
J.~Tremblay, T.~To, B.~Sundaralingam, Y.~Xiang, D.~Fox, and S.~Birchfield,
  ``Deep object pose estimation for semantic robotic grasping of household
  objects,'' \emph{arXiv preprint arXiv:1809.10790}, 2018.

\bibitem{huttenlocher1993comparing}
D.~P. Huttenlocher, G.~A. Klanderman, and W.~J. Rucklidge, ``Comparing images
  using the hausdorff distance,'' \emph{IEEE Transactions on pattern analysis
  and machine intelligence}, vol.~15, no.~9, pp. 850--863, 1993.

\bibitem{hinterstoisser2011gradient}
S.~Hinterstoisser, C.~Cagniart, S.~Ilic, P.~Sturm, N.~Navab, P.~Fua, and
  V.~Lepetit, ``Gradient response maps for real-time detection of textureless
  objects,'' \emph{IEEE transactions on pattern analysis and machine
  intelligence}, vol.~34, no.~5, pp. 876--888, 2011.

\bibitem{xiang2018posecnn}
Y.~Xiang, T.~Schmidt, V.~Narayanan, and D.~Fox, ``{PoseCNN}: A convolutional
  neural network for {6D} object pose estimation in cluttered scenes,'' in
  \emph{Robotics: Science and Systems (RSS)}, 2018.

\bibitem{wang2019densefusion}
C.~Wang, D.~Xu, Y.~Zhu, R.~Mart{\'\i}n-Mart{\'\i}n, C.~Lu, L.~Fei-Fei, and
  S.~Savarese, ``Densefusion: 6d object pose estimation by iterative dense
  fusion,'' in \emph{Proceedings of the IEEE/CVF conference on computer vision
  and pattern recognition}, 2019, pp. 3343--3352.

\bibitem{li2019cdpn}
Z.~Li, G.~Wang, and X.~Ji, ``Cdpn: Coordinates-based disentangled pose network
  for real-time rgb-based 6-dof object pose estimation,'' in \emph{Proceedings
  of the IEEE/CVF International Conference on Computer Vision}, 2019, pp.
  7678--7687.

\bibitem{peng2019pvnet}
S.~Peng, Y.~Liu, Q.~Huang, X.~Zhou, and H.~Bao, ``Pvnet: Pixel-wise voting
  network for 6dof pose estimation,'' in \emph{Proceedings of the IEEE/CVF
  Conference on Computer Vision and Pattern Recognition}, 2019, pp. 4561--4570.

\bibitem{he2020pvn3d}
Y.~He, W.~Sun, H.~Huang, J.~Liu, H.~Fan, and J.~Sun, ``Pvn3d: A deep point-wise
  3d keypoints voting network for 6dof pose estimation,'' in \emph{Proceedings
  of the IEEE/CVF conference on computer vision and pattern recognition}, 2020,
  pp. 11\,632--11\,641.

\bibitem{he2022fs6d}
Y.~He, Y.~Wang, H.~Fan, J.~Sun, and Q.~Chen, ``Fs6d: Few-shot 6d pose
  estimation of novel objects,'' in \emph{Proceedings of the IEEE/CVF
  Conference on Computer Vision and Pattern Recognition}, 2022, pp. 6814--6824.

\bibitem{liu2022gen6d}
Y.~Liu, Y.~Wen, S.~Peng, C.~Lin, X.~Long, T.~Komura, and W.~Wang, ``Gen6d:
  Generalizable model-free 6-dof object pose estimation from rgb images,''
  \emph{arXiv preprint arXiv:2204.10776}, 2022.

\bibitem{sun2022onepose}
J.~Sun, Z.~Wang, S.~Zhang, X.~He, H.~Zhao, G.~Zhang, and X.~Zhou, ``Onepose:
  One-shot object pose estimation without cad models,'' in \emph{Proceedings of
  the IEEE/CVF Conference on Computer Vision and Pattern Recognition}, 2022,
  pp. 6825--6834.

\bibitem{park2019deepsdf}
J.~J. Park, P.~Florence, J.~Straub, R.~Newcombe, and S.~Lovegrove, ``Deepsdf:
  Learning continuous signed distance functions for shape representation,'' in
  \emph{Proceedings of the IEEE/CVF conference on computer vision and pattern
  recognition}, 2019, pp. 165--174.

\bibitem{sitzmann2019scene}
V.~Sitzmann, M.~Zollh{\"o}fer, and G.~Wetzstein, ``Scene representation
  networks: Continuous 3d-structure-aware neural scene representations,''
  \emph{Advances in Neural Information Processing Systems}, vol.~32, 2019.

\bibitem{mescheder2019occupancy}
L.~Mescheder, M.~Oechsle, M.~Niemeyer, S.~Nowozin, and A.~Geiger, ``Occupancy
  networks: Learning 3d reconstruction in function space,'' in
  \emph{Proceedings of the IEEE/CVF conference on computer vision and pattern
  recognition}, 2019, pp. 4460--4470.

\bibitem{mildenhall2020nerf}
B.~Mildenhall, P.~P. Srinivasan, M.~Tancik, J.~T. Barron, R.~Ramamoorthi, and
  R.~Ng, ``Nerf: Representing scenes as neural radiance fields for view
  synthesis,'' in \emph{European conference on computer vision}.\hskip 1em plus
  0.5em minus 0.4em\relax Springer, 2020, pp. 405--421.

\bibitem{tang2020deep}
D.~Tang, S.~Singh, P.~A. Chou, C.~Hane, M.~Dou, S.~Fanello, J.~Taylor,
  P.~Davidson, O.~G. Guleryuz, Y.~Zhang, \emph{et~al.}, ``Deep implicit volume
  compression,'' in \emph{Proceedings of the IEEE/CVF conference on computer
  vision and pattern recognition}, 2020, pp. 1293--1303.

\bibitem{dupont2021coin}
E.~Dupont, A.~Goli{\'n}ski, M.~Alizadeh, Y.~W. Teh, and A.~Doucet, ``Coin:
  Compression with implicit neural representations,'' \emph{arXiv preprint
  arXiv:2103.03123}, 2021.

\bibitem{sitzmann2020implicit}
V.~Sitzmann, J.~Martel, A.~Bergman, D.~Lindell, and G.~Wetzstein, ``Implicit
  neural representations with periodic activation functions,'' \emph{Advances
  in Neural Information Processing Systems}, vol.~33, pp. 7462--7473, 2020.

\bibitem{adamkiewicz2022nerfnav}
\BIBentryALTinterwordspacing
M.~Adamkiewicz, T.~Chen, A.~Caccavale, R.~Gardner, P.~Culbertson, J.~Bohg, and
  M.~Schwager, ``Vision-only robot navigation in a neural radiance world,''
  \emph{CoRR}, vol. abs/2110.00168, 2021. [Online]. Available:
  \url{https://arxiv.org/abs/2110.00168}
\BIBentrySTDinterwordspacing

\bibitem{simeonov2022ndf}
\BIBentryALTinterwordspacing
A.~Simeonov, Y.~Du, A.~Tagliasacchi, J.~B. Tenenbaum, A.~Rodriguez, P.~Agrawal,
  and V.~Sitzmann, ``Neural descriptor fields: Se(3)-equivariant object
  representations for manipulation,'' \emph{CoRR}, vol. abs/2112.05124, 2021.
  [Online]. Available: \url{https://arxiv.org/abs/2112.05124}
\BIBentrySTDinterwordspacing

\bibitem{yen2021inerf}
L.~Yen-Chen, P.~Florence, J.~T. Barron, A.~Rodriguez, P.~Isola, and T.-Y. Lin,
  ``inerf: Inverting neural radiance fields for pose estimation,'' in
  \emph{2021 IEEE/RSJ International Conference on Intelligent Robots and
  Systems (IROS)}.\hskip 1em plus 0.5em minus 0.4em\relax IEEE, 2021, pp.
  1323--1330.

\bibitem{yu2021pixelnerf}
A.~Yu, V.~Ye, M.~Tancik, and A.~Kanazawa, ``pixelnerf: Neural radiance fields
  from one or few images,'' in \emph{Proceedings of the IEEE/CVF Conference on
  Computer Vision and Pattern Recognition}, 2021, pp. 4578--4587.

\bibitem{wang2019normalized}
H.~Wang, S.~Sridhar, J.~Huang, J.~Valentin, S.~Song, and L.~J. Guibas,
  ``Normalized object coordinate space for category-level 6d object pose and
  size estimation,'' in \emph{Proceedings of the IEEE/CVF Conference on
  Computer Vision and Pattern Recognition}, 2019, pp. 2642--2651.

\bibitem{ahmadyan2021objectron}
A.~Ahmadyan, L.~Zhang, A.~Ablavatski, J.~Wei, and M.~Grundmann, ``Objectron: A
  large scale dataset of object-centric videos in the wild with pose
  annotations,'' in \emph{Proceedings of the IEEE/CVF conference on computer
  vision and pattern recognition}, 2021, pp. 7822--7831.

\bibitem{hochreiter1997long}
S.~Hochreiter and J.~Schmidhuber, ``Long short-term memory,'' \emph{Neural
  computation}, vol.~9, no.~8, pp. 1735--1780, 1997.

\bibitem{kingma2014adam}
\BIBentryALTinterwordspacing
D.~P. Kingma and J.~Ba, ``Adam: {A} method for stochastic optimization,''
  \emph{CoRR}, vol. abs/1412.6980, 2014. [Online]. Available:
  \url{http://arxiv.org/abs/1412.6980}
\BIBentrySTDinterwordspacing

\bibitem{chang2015shapenet}
\BIBentryALTinterwordspacing
A.~X. Chang, T.~A. Funkhouser, L.~J. Guibas, P.~Hanrahan, Q.~Huang, Z.~Li,
  S.~Savarese, M.~Savva, S.~Song, H.~Su, J.~Xiao, L.~Yi, and F.~Yu, ``Shapenet:
  An information-rich 3d model repository,'' \emph{CoRR}, vol. abs/1512.03012,
  2015. [Online]. Available: \url{http://arxiv.org/abs/1512.03012}
\BIBentrySTDinterwordspacing

\bibitem{willis2022joinable}
K.~D. Willis, P.~K. Jayaraman, H.~Chu, Y.~Tian, Y.~Li, D.~Grandi, A.~Sanghi,
  L.~Tran, J.~G. Lambourne, A.~Solar-Lezama, and W.~Matusik, ``Joinable:
  Learning bottom-up assembly of parametric cad joints,'' in \emph{Proceedings
  of the IEEE/CVF Conference on Computer Vision and Pattern Recognition
  (CVPR)}, June 2022.

\bibitem{gmsd}
W.~Xue, L.~Zhang, X.~Mou, and A.~C. Bovik, ``Gradient magnitude similarity
  deviation: A highly efficient perceptual image quality index,'' in \emph{IEEE
  transactions on image processing, 23(2)}, 2013, p. 684–695.

\end{thebibliography}

\end{document}